\documentclass{article}
\usepackage{ijcai18}

\usepackage{times}
\usepackage{url}
\usepackage{xcolor}
\usepackage{soul}
\usepackage[utf8]{inputenc}
\usepackage[small]{caption}

\usepackage{amsmath}
\usepackage{amsfonts}
\usepackage{bbm}
\usepackage{tikz}
\usepackage{framed}

\newcommand{\fol}{\ensuremath{{\cal L}}}
\newcommand{\interp}{\ensuremath{{\cal I}}}
\newcommand{\evidence}{\ensuremath{e}}

\newcommand{\hbase}{\ensuremath{B_{{\cal L}}}}

\newcommand{\atom}{\ensuremath{\mathrm{Atom}}}

\newcommand{\cost}{\ensuremath{c}}
\newcommand{\atomvar}{\ensuremath{x(\mathrm{Atom})}}
\newcommand{\lpsol}{\ensuremath{x^*}}

\newcommand{\reals}{\ensuremath{\mathbb{R}}}
\newcommand{\realsnonneg}{\ensuremath{\mathbb{R}_{+}}}

\newcommand{\mfoilp}{{\sc mfoilp}}
\newcommand{\consfolinear}{{\tt folinear}}

\newcommand{\bob}{{\ensuremath{\mathrm{bob}}}}
\newcommand{\alice}{{\ensuremath{\mathrm{alice}}}}
\newcommand{\jim}{{\ensuremath{\mathrm{jim}}}}
\newcommand{\male}{{\ensuremath{\mathrm{male}}}}
\newcommand{\father}{{\ensuremath{\mathrm{father}}}}
\newcommand{\parent}{{\ensuremath{\mathrm{parent}}}}

\newcommand{\consfolinearc}{{\tt folinear.c}}
\newcommand{\cfoilpc}{{\tt cfoilp.c}}
\newcommand{\mfoilpm}{{\tt mfoilp.m}}
\newcommand{\scip}{{\tt SCIP}}
\newcommand{\cplex}{{\tt CPLEX}}
\newcommand{\probm}{{\tt prob.m}}

\newcommand{\rockit}{{\sc RockIt}}
\newcommand{\alchemy}{{\sc Alchemy}}
\newcommand{\tuffy}{{\sc Tuffy}}

\newcommand{\professor}{{\ensuremath{\mathrm{professor}}}}
\newcommand{\publication}{{\ensuremath{\mathrm{publication}}}}

\newcommand{\hasPosition}{{\ensuremath{\mathrm{hasPosition}}}}

\newcommand{\advisedBy}{{\ensuremath{\mathrm{advisedBy}}}}

\newcommand{\samePerson}{{\ensuremath{\mathrm{samePerson}}}}

\title{Finding Minimal Cost Herbrand Models with Branch-Cut-and-Price}
\author{James Cussens\\Dept of Computer Science, University of York, UK\\james.cussens@york.ac.uk}

\begin{document}

\maketitle

\begin{abstract}
  Given (1) a set of clauses $T$ in some first-order language \fol{}
  and (2) a cost function
  $\cost : \hbase \rightarrow \realsnonneg$, mapping each ground atom
  in the Herbrand base \hbase{} to a non-negative real, then the problem of
  finding a minimal cost Herbrand model is to either find a Herbrand
  model \interp{} of $T$ which is guaranteed to minimise the sum of
  the costs of true ground atoms, or establish that there is no
  Herbrand model for $T$. A branch-cut-and-price integer programming
  (IP) approach to solving this problem is presented. Since the number
  of ground instantiations of clauses and the size of the Herbrand base are
  both infinite in general, we add the corresponding IP constraints and
  IP variables `on the fly' via `cutting' and `pricing'
  respectively. In the special case of a finite Herbrand base we show
  that adding all IP variables and constraints from the outset can be
  advantageous, showing that a challenging Markov logic network MAP
  problem can be solved in this way if encoded appropriately.
  

\end{abstract}

\section{Introduction}
\label{sec:intro}

A (pure) integer programming (IP) problem consists of the following:
(1) a set of problem variables with upper and lower bounds, some of
which (perhaps all) are constrained to take integer values, (2) an
objective function which is a linear function of the variables and (3)
linear constraints on the variables. The goal of an IP solver is to
find a solution that satisfies all constraints and maximises (or
minimises) the
objective function. Decades of work on the theory, practice and
implementation of IP solvers means that they are often the method of
choice for solving hard (NP-hard) constrained optimisation problems.

IP problems are presented to IP solvers either via an API (the Gurobi
solver \cite{gurobi} has APIs for 6 programming languages) or using
some standard format or modelling language such as ZIMPL
\cite{Koch2004} or AMPL
\cite{fourer90:_model_languag_mathem_progr}. In the constraint
programming community MiniZinc \cite{nethercote07:_miniz} is used.
Modelling languages provide a `template' approach where the user can
declare many variables and constraints in a compact way. For example,
in the ZIMPL language
\begin{verbatim}
set I := { 1 .. 100 }; 
var x[I] integer >= 2 <= 18;
var y[I] real;
\end{verbatim}
declares 100 integer variables 100 real-valued variables and
\begin{verbatim}
subto fo: forall<i> in I: x[i] <= 2*y[i];
\end{verbatim}
declares 100 linear constraints. This compact representation is then
`unrolled' into an explicit list of variables and constraints, so that
the problem presented to the solver is exactly as if the user had
laboriously listed all variables and constraints explicitly in the
first place. This unrolling is done by either creating a large input
file (in say, `lp' format) or is done internally by the solver. For
example, the SCIP solver \cite{GleixnerEiflerGallyetal.2017} directly
accepts problems defined in the ZIMPL language.

In this paper we use first-order logic as the modelling language and
use it to describe problems which cannot, in general, be represented
by unrolling since this would result in infinitely many IP variables
and constraints. By way of introduction we first consider the special
case (first-order languages without function symbols) where a
first-order representation \emph{can} be unrolled.

\section{First-order logic as a template language}
\label{sec:template}

It is easy to use first-order logic to define a standard IP problem
with finitely many variables and constraints. We can write a logic
program (in e.g.\ Prolog) to define a predicate \texttt{linear/4}
using Prolog clauses like this:
\begin{verbatim}
linear(LHS,Coeffs,Var,RHS) :-
   foo(LHS,Coeffs,Var,RHS).

linear(LHS,Coeffs,Var,RHS) :-
   bar(LHS,Coeffs,Var,RHS).
\end{verbatim}
and the use Prolog's \verb+findall/3+ meta-predicate to find all ground
instances of \verb+linear(LHS,Coeffs,Var,RHS)+ which are logically
entailed by the logic program. (Throughout this paper logical
variables will have an initial upper case letter and constants,
function and predicate symbols will have an initial lower case
letter.) Each such ground instance will
represent a linear constraint in the IP. Other predicates can be
written to similarly generate IP variables together with their bounds
and objective coefficients. Many IP solvers go beyond pure IP and
allow non-linear (e.g.\ quadratic) constraints. Such constraints can
be defined analogously to the linear ones. Note that the variables and
constraints of the IP are represented as ground terms in the
problem-defining logic program.

\subsection{Encoding MLN MAP problems}
\label{sec:mlns}

Using first-order language as a template language is particularly
convenient when using IP to solve \emph{MAP problems for Markov logic
  networks (MLNs)}.  An MLN is a ``a finite set of pairs
$(F_{i},w_{i})$, $1 \leq i \leq n$, where each $F_i$ is a clause in
function-free first-order logic and $w_{i} \in \reals$.''
\cite{noessner13:_rockit}. Since the clauses are function-free the
number of ground atoms in the first-order language implicitly defined
by the MLN is finite. An MLN defines a probability for each Herbrand
interpretation \interp{} as follows:
$
  P(\interp) = \frac{1}{Z} \exp \left(\sum_{i=1}^{n} w_{i}
    n_{i}(\interp) \right)$
where $n_{i}(\interp)$ is the number of groundings of formula $F_i$
which are true in \interp{} and $Z$ is a normalising constant. The MAP
problem for MLNs is to find $\arg \max_{\interp} P(\interp|\evidence)$ where
\evidence{} is \emph{evidence}, a (possibly empty) set of ground atoms
with fixed truth values (true or false). In other words the goal is to
find the most probable interpretation (Herbrand model) which is consistent with the
evidence.

Similarly to the \rockit{} system \cite{noessner13:_rockit} we will
construct an IP where there is a one-one correspondence between
non-evidence ground atoms in the Herbrand base and binary variables in
the IP. We will present our encoding by example using the LP (`link
prediction') MLN which can be downloaded from the \tuffy{}
\cite{niu11:_tuffy} website.
The following weighted clause from the LP MLN:
$0.749 :  \neg \publication(A3,A1)
    \vee \neg
    \publication(A3,A2) \vee
     \samePerson \vee \advisedBy(A1,A2) \vee \advisedBy(A2,A1)$
is encoded as follows:
\begin{verbatim}
cons(lit(p,cb(19,A1,A2)),and,
     [lit(n,advisedBy(A1,A2)),
       lit(n,advisedBy(A2,A1))]) :-
    guard(19,A1,A2,[_A3]).

guard(19,A1,A2,[A3]) :-
 publication(A3,A1), publication(A3,A2),
 not samePerson(A1,A2).
\end{verbatim}
These 2 Prolog clauses says that for any grounding
$\{A1/a1,A2/a2,A3/a3\}$ which satisfies the \verb+guard/4+ predicate this 
constraint:
\begin{verbatim}
cb(19,a1,a2) <-> 
   ~advisedBy(a1,a2) & ~advisedBy(a2,a1)
\end{verbatim}
should be added to the IP. This constraint states that the binary variable
\verb+cb(19,a1,a2)+ is true (takes value 1) iff both
\verb+advisedBy(a1,a2)+ and \verb+advisedBy(a2,a1)+ are false. Both
the SCIP and Gurobi solvers accept AND constraints, like this. (SCIP internally
creates linear constraints from AND constraints, presumably Gurobi
does also.) In the LP MLN \verb+publication/2+ and \verb+samePerson/2+
are evidence predicates: the truth value of every one of their ground
instances are fixed. This is represented in our IP-defining Prolog
program by simply adding those \verb+publication/2+ facts which are
true. In the case of \verb+samePerson/2+ we add the single clause
\verb+samePerson(X,X)+
rather than adding the 68 facts like
\verb+samePerson("Person319", "Person319")+ which are present in the
original encoding of the problem.

Our encoding reflects the fact that we are only interested in
groundings of the MLN clause which are not satisfied due to evidence
ground atoms having a clause-satisfying truth value. The encoding
states that for each such grounding where \verb+advisedBy(a1,a2)+ and
\verb+advisedBy(a2,a1)+ have the `wrong' truth values (are false) then
\verb+cb(19,a1,a2)+ has to be true. \verb+cb(19,a1,a2)+ is a penalty
atom (\verb+cb+ stands for `clause broken') so must have a positive
cost. This means that in any optimal solution it will have value 1
only if forced to do so by the values of \verb+advisedBy(a1,a2)+ and
\verb+advisedBy(a2,a1)+. This is why we can have an AND (`iff')
constraint rather than a weaker `if' constraint. 

To get the correct
cost it is necessary to count how many $\{A1/a1,A2/a2,A3/a3\}$
groundings correspond to \verb+cb(19,a1,a2)+. This can be achieved as
follows:
\begin{verbatim}
cost(cb(19,A1,A2),Cost) :-
    setof(X,guard(19,A1,A2,X),Sols),
    length(Sols,Count),
    Cost is Count * 0.749123.
\end{verbatim}

The main features of our chosen encoding have now been
illustrated---most other MLN clauses are encoded just like the example
just given. Note that
when an MLN clause has only one non-evidence literal in it then the
resulting AND constraints are of the form \verb+cb <-> lit+,
i.e. equations. This allows pre-processing to remove the penalty atom
(and the AND constraint) from the problem.

The following MLN clause (clause 10) from the LP example:
$     0.385 :  \samePerson(A2,A3)
     \vee \neg
     \advisedBy(A1,A3) \vee
      \neg \advisedBy(A1,A2) 
$
cannot be so pre-processed away and will lead to many AND
constraints in the IP. However, since the predicate
\verb+samePerson/2+ is just equality in disguise it has a special
structure which can be exploited. This clause states that for any
\verb+a1+ there should be a penalty of 0.384788 for each ordered pair
\verb+(a2,a3)+ of distinct individuals where both
\verb+advisedBy(a1,a3)+ and \verb+advisedBy(a1,a2)+ are true. If $n$
is the number of facts (in a candidate model) which unify with
\verb+advisedBy(a1,_)+ then the number of such pairs is simply
$n^{2}-n$. So we encode clause 10 by one linear
constraint (to compute $n$) and one quadratic constraint (to compute
$n^{2}-n$).

The resulting IP contains 30,243 variables and 44,609 constraints. However,
after pre-processing we end up with only 7,108 variables and 2,484
constraints. Using the SCIP solver the IP is solved to optimality in
26.3 seconds (including 4.4 seconds for pre-processing) using a single
core of a 1.7GHz laptop. The optimal model found has 60 ground atoms
set to true. In contrast, as reported by Noessner \emph{et al}
\cite{noessner13:_rockit} none of the 4 MLN systems \rockit{}
\cite{noessner13:_rockit}, TheBeast \cite{Riedel08}, \tuffy{}
\cite{niu11:_tuffy} or \alchemy{}
\cite{kok07:_alchem_system_statis_relat_ai} were able to solve this
problem to optimality (or even to a 0.1\% gap) within 1 hour.

The key to solving this particular problem is dealing with clause 10
properly. If clause 10 is omitted both \rockit{} and our approach
solve the problem very quickly, both returning the same optimal
solution (with 273 ground atoms set to true). Using the version of
\rockit{} available via the \rockit{} web interface, with clause 10 included
\rockit{} does find an optimal solution (with 60 true ground atoms)
after 424 seconds but is unable to establish that it is optimal.

The version of the LP MLN considered so far contains 24 MLN clauses
\cite[Table~4]{noessner13:_rockit} and excludes 2 MLN formulae which
are weighted (i.e.\ not hard) formulae containing existentially
quantified variables. This is because \rockit{} cannot handle such
formulae. However, we can, for example encoding this formula:
$     -1.23183 : \exists Y  \neg \professor(X) \vee \advisedBy(Y,X) 
     \vee \hasPosition(X,``Faculty\_visiting") 
$
like this:
\begin{verbatim}
cons(lit(n,cb(26,X)),and,Lits) :-
  professor(X),
  \+ hasPosition(X,"Faculty_visiting"),
  findall(lit(n,advisedBy(Y,X)),person(Y)
          ,Lits).
\end{verbatim}
The IP resulting from adding in the two missing existentially
quantified formula is solved by SCIP in 29 seconds.

\section{The minimal cost Herbrand model problem}
\label{sec:findinf}

The basic idea of the current paper is that the template approach
described in Section~\ref{sec:template} can be extended to the case
where the logic program defines an IP with infinitely many variables
and constraints. Evidently, in this case, the simple `grounding out'
approach of Section~\ref{sec:template} can no longer be used. Instead
variables and constraints of the IP will be added to the IP during the
course of solving by methods known as `pricing' and `cutting'
respectively. We show that in some cases only a finite subset of the
full set of variables and constraints need be added to find a provably
optimal solution.

We will initially restrict attention to problems where each constraint
in the underlying IP is a linear constraint corresponding to a ground
clause in some first-order language and these (perhaps infinitely
many) ground clauses are defined as the set of all ground instances of
some finite collection of first-order clauses. Formally, we consider
the problem of finding a minimal cost Herbrand model: given (1) a set
of clauses $T$ in some first-order language \fol{} and (2) a
\emph{cost function} $\cost : \hbase \rightarrow \realsnonneg$,
mapping each ground atom in the Herbrand base to a non-negative real,
then our goal is either to find a Herbrand model \interp{} of $T$
which is guaranteed to minimise the sum of the costs of true ground
atoms, or to establish that there is no Herbrand model for $T$.

The IP encoding of the problem is straightforward.  For each ground
atom $\atom \in \hbase$ there is a binary IP variable \atomvar{} with
objective coefficient $\cost(\atom)$. Only a finite subset of these IP
variables ever get explicitly represented in the IP. There is also a
linear inequality for each grounding of each of clauses in $T$. For
a ground clause:
$\neg \atom_{1} \vee \dots \vee \neg \atom_{r'} \vee \atom_{r'+1} \vee
\dots \vee \atom_{r}$ the corresponding linear inequality is:
\begin{equation}
  \label{eq:clauseineq}
  \begin{split}
  [1-x(\atom_{1})] &+ \dots +  [1-x(\atom_{r'})]  \\
  + x(\atom_{r'+1}) &+ \dots +  x(\atom_{r}))  \geq 1
\end{split}
\end{equation}
Call such inequalities \emph{clausal inequalities}
\cite{hooker07:_integ_method_optim}. Of course, only a finite subset
of the clausal inequalities are ever present in the IP.

\section{Cut-and-price}
\label{sec:cutprice}

Consider first the special case where the first-order language \fol{}
contains no function symbols apart from constants, so \hbase{} and
thus the number of IP variables and constraints are both finite. Let
$n$ denote the number of variables, and let
$x \in \{0,1\}^n$ denote a generic candidate solution to the IP.
Typically, the strategy for solving such an IP would involve solving
the \emph{linear relaxation} of the IP which provides a useful global
lower bound on any optimal solution. The linear relaxation is the
linear program (LP) which results from relaxing the integrality
constraint $x \in \{0,1\}^n$ to $x \in [0,1]^n$.  However, if there
were very many ground clauses (= very many clausal inequalities) then
solving this LP (which we call the \emph{full LP}) could be slow. A
\emph{cutting plane} approach addresses this problem: a small number
(perhaps zero) of linear inequalities from the full LP are included in
an initial LP. This initial LP is solved producing a solution
$x^*$. There is then a search for one or more linear inequalities from
the full LP which $x^*$ does not satisfy; such inequalities are known
as cutting planes or \emph{cuts}. If any cuts are found they are added
to the initial LP which is then re-solved, generating a new solution
$x^*$. This process continues until no cuts can be found, at which
point we have an optimal solution to the full LP even though
(typically) we have not added all its linear inequalities. This
approach can be strengthened by not only using inequalities from the
full LP but additional inequalities which follow from assuming that
$x$ is integer-valued. Adding such additional inequalities results in
a LP whose solution provides a better lower bound on the IP solution.

It may be that the number of \emph{variables} in the full LP (i.e.\
the size of the Herbrand base) is also so big as to cause problems. To
get round this problem one can adopt a \emph{pricing} strategy, where
in an initial LP only a small number (perhaps zero) of variables are used. All
omitted variables are implicitly fixed to have value zero, so if an
omitted variable appears in a clausal inequality it is replaced with a
zero. Once this initial LP is solved there is then a search for
currently omitted variables which, if allowed to take a value other
than zero, would allow a better solution to the current LP (better
either in terms of feasibility or objective value). If the current LP
has a solution then any variable with \emph{negative reduced cost}
allows an improvement. Reduced costs are now explained.  If the
current LP has a solution $x^*$ then there will also be a solution,
call it $\lambda^{*} \in \realsnonneg^{m}$, to its \emph{dual}, which
assigns a non-negative real value $\lambda^{*}_{j}$ ($j=1,\dots,m$) to
each of the $m$ inequalities in the LP. Let $a_{\atomvar,j}$ be the
coefficient of variable \atomvar{} in inequality $j$, then the
\emph{reduced cost} for \atomvar{} associated with solution $x^*$ is
$\cost(\atomvar) - \sum_{j=1}^{m} \lambda^{*}_{j} a_{\atomvar,j}$. (If
the current LP is infeasible then there will be a vector $\lambda^{*}
\in \realsnonneg^{m}$ of \emph{dual Farkas multipliers} which allows
`improving' variables to be identified in a similar way, a process
called \emph{Farkas pricing}.)
If the current LP has a solution and no omitted variable with negative
reduced cost can be found then it follows that the current set of
variables is enough to get an optimal solution to the LP.

The \emph{cut-and-price} approach used in this paper rests on the
simple observation that both cutting and pricing can still be used
when the `pool' of available inequalities and variables are allowed to
be infinite, rather than just very large but finite. This means that
the above described price-and-cut approach can be applied when there
is no restriction on the first-order language. So from now on we will
remove the restriction to finite Herbrand bases and consider general
first-order languages.

\section{Generating cuts from first-order clauses}
\label{sec:gencuts}

We now consider how to find cutting planes, a problem known as the
\emph{separation problem}. Let $T$ be a set of first-order clauses (a
CNF formula). For the time being we will make the
simplifying assumption that any substitution that grounds all the
negative literals in a first-order clause in $T$ determines a 
grounding for all positive literals in the clause. We will later
consider how this restriction can be relaxed.

Given a solution $x^*$ to an LP whose inequalities are a
finite set of ground instances of these clauses, the problem is to
find a new ground instance of some first-order clause in $T$ which
$x^*$ does not satisfy. This is done by considering each first-order
clause in turn and for each doing a simple depth-first search for a
suitable ground instance. 
Each state of this search is a 4-tuple $(\theta,z,n,p)$ where
$\theta$ is a substitution, $n$ is a
set of ground atoms representing negative literals, $p$ a set of
ground positive literals and $z \in \realsnonneg$ is an
\emph{activity} value equal to
$\sum_{\atom \in n}[1-x^{*}(\atom)] + \sum_{\atom \in
  p}[x^{*}(\atom)]$. A state is a goal state if (i) $\theta$ is a 
complete grounding of the first-order clause and (ii) $z<1$.  The initial
state is $(\emptyset, 0,\emptyset,\emptyset)$. We first illustrate the
search process by example, and then describe it formally.  Consider
the following clause
\[
\neg \male(X) \vee \neg \parent(X,Y) \vee
\father(X,Y)
\]
and LP solution $x^*$ where $x^{*}(\male(\bob)) =0.4$,
$x^{*}(\parent(\bob,\jim)) =0.5$ and 
$x^{*}(\parent(\bob,\alice)) =0.9$.

The search grounds a given negative literal by scanning the LP solution $x^*$
for atoms such that $x^{*}(\atom) > 0$ and which unify with the
negative literal. In this example, we have $x^{*}(\male(\bob))=0.4$,
so we can ground the first literal using the substitution $\{X/\bob\}$
and increase the activity value to 1-0.4 = 0.6, allowing the search to
move to state:
$(\{X/\bob\},0.6,\{\male(\bob)\},\emptyset)$.

Suppose next that the search uses the fact that
$x^{*}(\parent(\bob,\jim))=0.5$ to unify $Y$ with $\jim$. This leads
to state:
$(\{X/\bob,Y/\jim\},1.1,\{\male(\bob),\parent(\bob,\jim)\},\emptyset)$. This
is a fail-state since $1.1 > 1$ and so the search would
backtrack and use $x^{*}(\parent(\bob,\alice)=0.9$ to unify $Y$ with $\alice$, leading to
state:
$(\{X/\bob,Y/\alice\},0.7,\{\male(\bob),\parent(\bob,\alice)\},\emptyset)$.

Both negative literals are now ground and since the clause satisfies
the restriction mentioned above, the positive literal is also ground
and is the atom $x(\father(\bob,\alice))$.
Suppose now that
$x(\father(\bob,\alice))$ is an omitted variable so
its value is zero in \lpsol{}. In this case we have reached the goal state:
$  (\{X/\bob,Y/\alice\},0.7,\{\male(\bob),\parent(\bob,\alice)\}$,
  $\{\father(\bob,\alice\})$
which corresponds to the cut
$[1-x(\male(\bob))] + [1-x(\parent(\bob,\alice))]  \geq 1$.
Note that this inequality:
$[1-x(\male(\bob))] + [1-x(\parent(\bob,\alice))]  + 
x(\father(\bob,\alice)) \geq 1$
could only be generated if the currently-omitted variable
$x(\father(\bob,\alice))$ were created---an issue discussed in the
next section.

The search for cuts from first-order clauses is now formally described. We
assume that all negative literals precede positive literals in each
first-order clause and that any grounding of all negative literals
determines a unique grounding for all positive literals.  If the
current state is $(\theta,z,n,p)$ and the next literal in the
first-order clause is $\neg \atom$ then successor states are of the
form
$(\theta\theta',z+[1-x^{*}(\atom\theta\theta')],n\cup\{\atom\theta\theta'\},p)$
where $\atom\theta\theta'$ is a grounding of $\atom$ such that
$x^{*}(\atom\theta\theta') > 0$. If the next literal is a positive
literal $\atom$ then $\atom\theta$ will be ground, and the unique
successor state is
$(\theta,z+x^{*}(\atom\theta),n,p\cup\{\atom\theta\})$ where
$x^{*}(\atom\theta)$ is defined to be zero if the LP variable
$x^{*}(\atom\theta)$ does not currently exist.  

Note that this search will always terminate since the following are
all finite: (1) the number of atoms \atom{} such that $x^*(\atom)>0$,
(2) the number of first-order clauses and (3) the length of each
first-order clause. Moreover, for similar reasons the number of cuts
which can be generated by this search is also finite.

\section{Generating ground atoms}
\label{sec:genvars}

In a typical cut-and-price approach, \emph{pricing}---the search for
improving variables---is done separately from cutting, and indeed,
earlier (unpublished) work of ours took just
this approach. However, it is possible to efficiently generate
improving variables as part of cut generation, avoiding the need for a
separate search. This is the approach taken here.

Recall that the reduced cost of a variable is
$\cost(\atomvar) - \sum_{j=1}^{m} \lambda^{*}_{j} a_{\atomvar,j}$ from
which it immediately follows that a variable can only have negative
reduced cost if its associated ground atom appears as a positive
literal in at least one of the $m$ ground clauses. (The only `reason'
for setting a ground atom to true is if doing so `helps' satisfy a
(ground instance of) a clause.) This leads to the following very
simple pricing strategy: for each clausal inequality in the current LP
ensure that all variables corresponding to positive literals exist in
the LP (creating them if necessary). A drawback of this approach is
that it may lead to the creation of more LP variables than
necessary---since we create all LP variables which might conceivably have
negative reduced cost rather than search for those that definitely
do. The simplicity and efficiency of creating new inequalities and new
variables simultaneously is however sufficient compensation.

So, in the example cut given in Section~\ref{sec:gencuts} we would
create the missing variable
$x(\father(\bob,\jim))$ and add it to the
generated cut giving
$[1-x(\male(\bob))] +
[1-x(\parent(\bob,\alice))] +  x(\father(\bob,\jim))  \geq 1$
This is still a cut for the current LP solution $x^*$ since
$x^{*}(\father(\bob,\jim))=0$. However now
that $x(\father(\bob,\jim))$ exists in the LP
a better solution where
$x(\father(\bob,\jim))$ has a positive value
may be possible.


\section{Branch-price-and-cut}
\label{sec:bpc}

Our approach is to create an initial IP with no variables and no
inequalities and to search for cuts (for the IP's linear relaxation)
using the method given in Section~\ref{sec:gencuts}, adding variables
at the same time, as described in Section~\ref{sec:genvars}. Assuming
at least one cut is found this produces a new linear relaxation for
which cuts are sought (and new variables generated) in the same
way. This process continues until no further cuts can be
found. However, we have no guarantee that this will terminate since
the problem of determining whether a set of first-order clauses even
has a model is undecidable. In practice, we impose a time limit and
admit defeat if it is reached before the problem is solved.

If the cut-generating process terminates, the objective value of the
solution ($x^*$) to this final LP provides a global lower bound on
solutions to the IP. If $x^*$ happens to be an integer solution then
the IP is solved. However, typically this is not the case and there
will be \emph{fractional} values $x^{*}(\atom)$ where
$0 < x^{*}(\atom) < 1$. If this is the case we \emph{branch} on some
fractional variable creating two subproblems, one where $x(\atom)=0$
(\atom{} is false) and one where $x(\atom)=1$ (\atom{} is true). The
solving process then continues recursively: each subproblem is
attacked in the same way as the original global problem. This method
of solving IPs is known as \emph{branch-price-and-cut}.

\section{Defining a problem instance}
\label{sec:probdefinition}

A problem instance is a triple $(\fol,T,\cost)$--- a first-order
language, a set of clauses and a cost function---and is defined by
writing a logic program in Mercury. Mercury \cite{zoltan96:_mercur} is
a strongly-typed, purely declarative logic programming language where
the user is obliged to declare types, modes and determinisms for each
predicate definition in a logic program. This allows Mercury programs
to be compiled to C and thence to native code. The result is much
faster execution than Prolog.

To define a problem instance the user must first define the first-order
language for the instance. This is done via a Mercury type declaration
specifying the Herbrand base \hbase. For example, this type
declaration:
\begin{verbatim}
:- type atom ---> 
     f(int,list(int)) ; cb(int,list(int)).
\end{verbatim}
declares that \hbase{} includes \verb+f(3,[2,4])+,
\verb+f(9,[2,4,4])+, \verb+cb(2,[4])+, \verb+cb(200,[5,6])+, and all other
(infinitely many) similarly typed ground atoms. (Note that one can
view integers as abbreviations for ground terms in some suitable
first-order language, where, for example, ``2'' abbreviates
``s(s(0))''.)

Secondly, the user is required to declare the cost function by
defining a \emph{semi-deterministic} predicate mapping atoms to
floats. Continuing our example, we might have:
\begin{verbatim}
:- pred cost(atom::in,
             float::out) is semidet.
cost(cb(X,L),1.0/float(X)).
cost(f(X,[H|T]),0.01).
\end{verbatim}
The ``semidet'' (i.e.\ semi-deterministic) declaration states that
\texttt{cost/2} either maps an input ground atom to a unique float or
fails.  Note that ground atoms in \hbase{} are ground \emph{terms} in
the Mercury program, so that the problem-defining Mercury program is, in effect, a
meta-program.  Allowing failure in \texttt{cost/2} is just for
convenience since this relieves the user from having to explicitly
define zero costs: a ground atom for which \texttt{cost/2} fails
implicitly has zero cost. In this particular case, \texttt{cost/2}
always succeeds so the Mercury compiler will internally convert
\texttt{cost/2} into a function.

Thirdly, the user must represent the first-order clauses. Continuing our
example suppose this clause were in $T$: 
$\forall N,L :  \neg f(N,L) \vee f(N+1,[N+1|L]) \vee cb(N,L)$,  
then it would be represented in the problem-defining Mercury
program as follows:
\begin{verbatim}
clause("2") -->
 neglit_out(f(N,L)), neglit(f(N,L)),
 poslit(f(N+1,[N+1|L])), poslit(cb(N,L)).
\end{verbatim}
Here DCG-notation is being used, so each of the 5 literals in the
Mercury clause has 2 extra variables which are not explicitly
represented. These extra variables represent states of the search for
cuts which was described in Section~\ref{sec:gencuts}. \verb+"2"+ is
just an arbitrary identifier for the clause. Note that there are 2
Mercury predicates for the negative literal $f(N,L)$. The first,
\verb+neglit_out/3+, generates a grounding from the current LP solution
and the second, \verb+neglit/3+, does everything else that is
necessary. This division of labour is for reasons of efficiency and
could perhaps be hidden from the user by some syntactic sugar.

Given an LP solution \lpsol{}, the Mercury goal
\verb+clause("2",S0,S4)+ is called where \verb+S0+ will be unified
with a ground term representing the initial state of the search. If
this goal succeeds then \verb+S4+ will represent a goal state of the
search from which a cut can be extracted and added to the LP. Using
Mercury's builtin-in \verb+solutions/2+ predicate (Mercury's version
of Prolog's \verb+findall/3+) we can find all valid instantiations of
\verb+S4+ and thus all groundings of the clause which are cuts for
\lpsol.

\subsection{Using context predicates}
\label{sec:context}

This Mercury clause
\begin{verbatim}
clause("walls") -->
    neglit_out(position(I,X1,Y1)),
    neglit_out(position(I+1,X2,Y2)),
    {wall_between(I,X1,Y1,X2,Y2)},
    neglit(position(I,X1,Y1)),
    neglit(position(I+1,X2,Y2)).
\end{verbatim}
generates ground instances of the clause
$\neg \mathrm{position(I,X1,Y1)} \vee \neg
\mathrm{position(I+1,X2,Y2)}$
but only those where \verb+wall_between(I,X1,Y1,X2,Y2)+ is true. (The
curly brackets indicates that no extra `state' arguments are added.)
\verb+wall_between/5+ is a \emph{context predicate} whose definition
is given by a normal Mercury clause, for example:
\verb|wall_between(I,X,Y,X+1,Y) :- I mod 3 = 0.|

The set of true ground atoms for a context predicate are fixed (by its
definition in the problem-defining Mercury program) before solving
even begins. Such atoms are implicitly in \hbase{} but since their
truth values are given it would be inefficient to represent them by IP
variables. Context predicates play a similar role to evidence
predicates in MLNs.

In the first paragraph of Section~\ref{sec:gencuts} we promised to
remove the restriction that the negative literals in a clause must
determine a grounding for the positive literals. 
The use of context predicates allows the removal of this restriction
since they can be used instead to generate the required grounding.

\section{Implementation}
\label{sec:implementation}

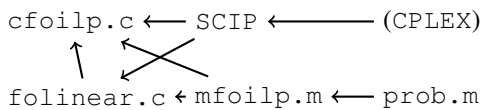
\begin{figure}
  \centering
   \begin{tikzpicture}[xscale=2.5]
      \node[anchor=west] (consfolinear) at (0,0) {\consfolinearc};
      \node[anchor=west] (cfoilp) at (0,1) {\cfoilpc};
      \node[anchor=west] (mfoilp) at (1,0) {\mfoilpm};
      \node[anchor=west] (scip) at (1,1) {\scip};
      \node[anchor=west] (prob) at (2,0) {\probm};
      \node[anchor=west] (cplex) at (2,1) {(\cplex)};
      \draw[->,thick] (consfolinear) -- (cfoilp);
      \draw[->,thick] (scip) -- (cfoilp);
      \draw[->,thick] (mfoilp) -- (cfoilp);
      \draw[->,thick] (mfoilp) -- (consfolinear);
      \draw[->,thick] (scip) -- (consfolinear);
      \draw[->,thick] (prob) -- (mfoilp);
      \draw[->,thick] (cplex) -- (scip);
    \end{tikzpicture}
  \caption{\mfoilp{} implementation. CPLEX, if available, used by SCIP for solving LPs}
  \label{fig:mfoilpimplementation}
\end{figure}

Our branch-price-and-cut algorithm for finding minimal cost Herbrand
models is called \mfoilp{}
(\url{https://bitbucket.org/jamescussens/mfoilp/}) and is implemented
in C and Mercury using the SCIP Optimization Suite
\cite{GleixnerEiflerGallyetal.2017}. Fig~\ref{fig:mfoilpimplementation}
shows how \mfoilp{} is organised.

\mfoilp{} is essentially the SCIP solver equipped with an extra
\emph{constraint handler} called \consfolinear{} which handles
constraints which are first-order clauses. Just like the 30 constraint
handlers already included in the current version of SCIP,
\consfolinear{} provides callbacks for checking whether candidate
solutions meet constraints, generating cuts, etc.

The problem instance is defined by Mercury predicate definitions
in the Mercury program \probm{}. \probm{} must be compiled before solving
begins. Once object code for \probm{} has been generated it is then
linked with (already-generated) object code for the rest of \mfoilp{}
thus generating a problem instance-specific executable which is then
executed to solve the problem. A Makefile is used to keep track of
what, if anything, needs recompiling before solving begins.

\section{Using \mfoilp}
\label{sec:experiments}

Whether a given problem with first-order clausal constraints is
solvable, and if so whether reasonably quickly, is largely determined
by the problem at hand.

We have tested \mfoilp{} on a number of problems where \hbase{} is
infinite. We have checked that when each clause in $T$ has a negative
literal then \mfoilp{} immediately deduces that setting all atoms in
\hbase{} to false is an optimal solution. If each clause in $T$ is
definite (has exactly one positive literal) then \mfoilp{} generates
the minimal model for $T$, familiar from logic programming theory,
irrespective of the cost function. (Of course, this generation does
not terminate if the minimal model is infinite!)

We have created an infinite maze problem where (i) an agent has to
keep moving (to an adjacent location) until it reaches a goal
location, (ii) where walls appear and disappear dynamically (see the
clause in Section~\ref{sec:context}), and (iii) where each move has
unit cost.  We did not include a clause stating that the agent must
stop once it reaches a goal state, leaving \mfoilp{} to deduce that to
keep moving would be suboptimal.

Defining a goal location thus: \verb+goal(X,Y):-X > 1,Y > 4+ and
stating that the agent must be at square (0,0) at time point 0,
\mfoilp{} finds a minimal cost route of 7 steps to the goal location
(2,5) in 5.09 seconds using a single core of a 1.7GHz
laptop. \mfoilp{} generates over 14,000 ground clauses but only 305
ground atoms. The branching in our branch-price-and-cut algorithm
created 4889 nodes in the search tree.  The 14,000 ground clauses are
not distinct since, at present, we allow SCIP to remove `old' cuts
which are not tight for the current linear relaxation solution. This
keeps the size of each LP small---the largest one had only 140
constraints---but means that discarded cuts might need to be re-found
later on.

We have also created variants of this problem where the problem was
not solved within a 30 minute cutoff.  Generally, in our `maze'
experiments we have observed that if \mfoilp{} can find an optimal
solution it can quickly prove that it is optimal, but in other cases
no feasible solution can be found (not even suboptimal ones). At
present \mfoilp{} relies on SCIP's default \emph{primal heuristics} to
generate candidate solutions. In our maze experiments SCIP's
`simplerounding' algorithm, which generates integer solutions from LP
relaxation solutions, was what produced candidate solutions when \mfoilp{}
succeeded.
We expect that it would be beneficial to add to \mfoilp{} a
specialised primal heuristic which generates candidate Herbrand
models. Some variant of the standard method for generating the minimal
model of a definite program would be worth exploring.

\section{Conclusions and future work}
\label{sec:conclusions}

In this paper we have presented methods for integrating first-order
logical inference into integer programming, focusing on the problem of
finding a minimal cost Herbrand model. Note that the MAP problem for
MLNs is a special case of this problem. We hope that this paper will
stimulate further work in this direction, since much remains to be
done.

Most importantly, automatic reformulation of IP problems posed in
terms of (first-order) clauses is needed. Representing each
(ground) clausal constraint by its corresponding clausal inequality
(\ref{eq:clauseineq}), as \mfoilp{} does, is known to be a poor IP
formulation since it leads to a weak LP relaxation. This issue has
been analysed in some depth (for propositional logic) by Hooker
\cite{hooker07:_integ_method_optim} who provides the following
example. Given a CNF with these 4 clauses
$x_{1} \vee x_{2} \vee x_{3}$, $x_{1} \vee x_{4}$, $x_{2} \vee x_{4}$
and $x_{3} \vee x_{4}$, the best formulation (the `convex hull'
formulation) is not the corresponding 4 clausal inequalities but this
single inequality: $x_{1} + x_{2} + x_{3} + 2x_{4} \geq 3$. Hooker
also shows how adding clausal inequalities which are produced by
 \emph{resolution} on initially-given (propositional) clauses can
tighten the linear relaxation. Applying first-order resolution on
first-order clauses is thus particularly attractive since it amounts to
doing very many propositional resolutions in one step.
 
The big win achieved by reformulating the link prediction (LP) MLN
MAP problem (see Section~\ref{sec:mlns}) is thus just one example of a
general phenomenon. Our expectation is that having an initial
representation in first-order logic will make it easier for a problem
to be automatically transformed into a better formulation.

\bibliographystyle{plain}
\bibliography{submit}

\begin{thebibliography}{10}

\bibitem{fourer90:_model_languag_mathem_progr}
Robert Fourer, David~M. Gay, and Brian~W. Kernighan.
\newblock A modeling language for mathematical programming.
\newblock {\em Management Science}, 36:519--554, 1990.

\bibitem{GleixnerEiflerGallyetal.2017}
Ambros Gleixner, Leon Eifler, Tristan Gally, Gerald Gamrath, Patrick Gemander,
  Robert~Lion Gottwald, Gregor Hendel, Christopher Hojny, Thorsten Koch,
  Matthias Miltenberger, Benjamin M{\"u}ller, Marc~E. Pfetsch, Christian
  Puchert, Daniel Rehfeldt, Franziska Schl{\"o}sser, Felipe Serrano, Yuji
  Shinano, Jan~Merlin Viernickel, Stefan Vigerske, Dieter Weninger, Jonas~T.
  Witt, and Jakob Witzig.
\newblock The {SCIP Optimization Suite} 5.0.
\newblock Technical Report 17-61, ZIB, Takustr.7, 14195 Berlin, 2017.

\bibitem{hooker07:_integ_method_optim}
John~H. Hooker.
\newblock {\em Integrated Methods for Optimization}.
\newblock Springer, 2007.

\bibitem{gurobi}
Gurobi~Optimization Inc.
\newblock Gurobi optimizer reference manual, 2016.

\bibitem{Koch2004}
Thorsten Koch.
\newblock {\em Rapid Mathematical Programming}.
\newblock PhD thesis, Technische {Universit\"at} Berlin, 2004.
\newblock ZIB-Report 04-58.

\bibitem{kok07:_alchem_system_statis_relat_ai}
Stanley Kok, Parag Singla, Matthew Richardson, Pedro Domingos, Marc Sumner, and
  Hoifung Poon.
\newblock {\em The Alchemy System for Statistical Relational AI: User Manual}.
\newblock University of Washington, 2007.

\bibitem{nethercote07:_miniz}
N.~Nethercote, P.~J. Stuckey, R.~Becket, S.~Brand, G.~J. Duck, and G.~Tack.
\newblock {MiniZinc}: Towards a standard {CP} modelling language.
\newblock In C.~Bessiere, editor, {\em Proceedings of the 13th International
  Conference on Principles and Practice of Constraint Programming}, volume 4741
  of {\em {LNCS}}, pages 529--543. Springer, 2007.

\bibitem{niu11:_tuffy}
F.~Niu, C.~R\'{e}, A.~Doan, and J.~Shavlik.
\newblock Tuffy: Scaling up statistical inference in {M}arkov logic networks
  using an {RDBMS}.
\newblock In {\em Proceedings of the VLDB Endowment}, volume~4, pages 373--384,
  2011.

\bibitem{noessner13:_rockit}
Jan Noessner, Mathias Niepert, and Heiner Stuckenschmidt.
\newblock Rockit: Exploiting parallelism and symmetry for {MAP} inference in
  statistical relational models.
\newblock In {\em Proceedings of the Twenty-Seventh {AAAI} Conference on
  Artificial Intelligence, July 14-18, 2013, Bellevue, Washington, {USA.}},
  2013.

\bibitem{Riedel08}
Sebastian Riedel.
\newblock Improving the accuracy and efficiency of {MAP} inference for {Markov}
  logic.
\newblock In {\em Proceedings of the Twenty-Fourth Conference Annual Conference
  on Uncertainty in Artificial Intelligence (UAI-08)}, pages 468--475,
  Corvallis, Oregon, 2008. AUAI Press.

\bibitem{zoltan96:_mercur}
Zoltan Somogyi, Fergus Henderson, and Thomas Conway.
\newblock The execution algorithm of {M}ercury: an efficient purely declarative
  logic programming language.
\newblock {\em Journal of Logic Programming}, 29(1--3):17--64, October-December
  1996.

\end{thebibliography}

\end{document}